\documentclass{article}

\usepackage[final,nonatbib]{infocog_neurips_2022}

\usepackage[utf8]{inputenc} 
\usepackage[T1]{fontenc}    
\usepackage{hyperref}       
\usepackage{url}            
\usepackage{booktabs}       
\usepackage{amsfonts}       
\usepackage{nicefrac}       
\usepackage{microtype}      
\usepackage{xcolor}         

\usepackage{algorithm}
\usepackage{algpseudocode}
\usepackage{amsmath}
\usepackage{graphicx}
\usepackage{subcaption}
\usepackage{wrapfig}

\let\vec\mathbf

\usepackage[%
backend=biber,%
natbib=true,%
style=apa,%
uniquename=false,%
useprefix=false%
]{biblatex}
\title{On Narrative Information and the Distillation of Stories}

\author{%
    Dylan R.~Ashley $^{1,2,3}$
    \thanks{Equal contribution. Correspondence to \texttt{dylan.ashley@idsia.ch} and \texttt{vincent.herrmann@idsia.ch}}
    \And
    Vincent Herrmann $^{1,2,3}$ $^{*}$
    \AND
    Zachary Friggstad $^{4}$
    \And
    J{\"{u}}rgen Schmidhuber $^{1,2,3,5,6}$
    \AND
    $^1$ \normalfont{Dalle Molle Institute for Artificial Intelligence Research, Lugano, Switzerland} \\
    $^2$ \normalfont{Universit{\`{a}} della Svizzera italiana, Lugano, Switzerland} \\
    $^3$ \normalfont{Scuola universitaria professionale della Svizzera italiana, Lugano, Switzerland} \\
    $^4$ \normalfont{University of Alberta, Edmonton, Canad}a\\
    $^5$ \normalfont{NNAISENSE, Lugano, Switzerland} \\
    $^6$ \normalfont{AI Initiative, King Abdullah University of Science and Technology, Thuwal, Saudi Arabia}
}

\begin{document}

\maketitle

\begin{abstract}

The act of telling stories is a fundamental part of what it means to be human.
This work introduces the concept of narrative information, which we define to be the overlap in information space between a story and the items that compose the story.
Using contrastive learning methods, we show how modern artificial neural networks can be leveraged to distill stories and extract a representation of the narrative information.
We then demonstrate how evolutionary algorithms can leverage this to extract a set of narrative templates and how these templates---in tandem with a novel curve-fitting algorithm we introduce---can reorder music albums to automatically induce stories in them.
In the process of doing so, we give strong statistical evidence that these narrative information templates are present in existing albums.
While we experiment only with music albums here, the premises of our work extend to any form of (largely) independent media.

\end{abstract}

\section{Introduction}
\label{sec:introduction}

Our ability to comprehend and devise narratives is a fundamental aspect of our nature (see, e.g., \citet{ricoeur1991narrative}), so much so that some dub humans the storytelling animal~\citep{gottschall2012storytelling}.
Furthermore, work in cognitive psychology often relates our ability for story comprehension to the theory of mind, showing an overlap in their cognitive networks~\citep{mar2011neural}.
This overlap suggests that one of the reasons we engage in storytelling is to hone our social cognition.

However, some aspects of stories go beyond the interaction and interpretation of cognitive agents---whether they be fiction or real.
Stories are embedded even in collections of largely independent media that do not explicitly feature cognitive agents.
For instance, we can talk about narrative choices curators make when arranging an art exhibition or the overarching story that the tracklist of a concert induces.
This work investigates a very general notion of \textit{stories} as being collections of \textit{atoms} (i.e., words, images, etc.) and some meaningful ordering over the atoms, which we refer to as the \textit{narrative}.
In line with this, we define here the \textit{narrative information} of an atom as the mutual information between the atom and the story itself.
In Section~\ref{sec:narrative_essence}, we define the \textit{narrative essence} of an atom as the low-dimensional learned representation of its narrative information and show how contrastive learning can be used to obtain it.
In Section~\ref{sec:experiments_on_music_albums}, we experimentally show that narrative essence can be used to extract prototypical narrative templates from music albums that partially explain the order in which the songs are arranged.
We believe that qualitatively similar results should hold for most collections of (largely) independent media.

\section{Narrative Essence}
\label{sec:narrative_essence}

A strong narrative can be induced into a media collection simply by putting it in a specific order.
In doing so, we can choose to begin our story with a bang or slowly build up excitement; we can place a climax at a particular position and then end the story on a high or a low note.
There are intrinsic properties of each atom that determine its function and, thus, placement, in a narrative.
This insight leads to our formulation of \textit{narrative essence}: a low-dimensional representation of the latent property of the items that is most informative about the narrative.

Formally, we define as narrative essence $f_E(x)$ of atom $x$, generated by a feature extractor $f_E$, as the feature which maximizes the mutual information between the unordered set of features of the atoms in a collection $c$ and the ground truth order $o(c)$ of $c$: \(f_E = \arg \max_{f} I\big(\{f(x) | x \in c\} ; o(c)\big)\).
In other words, narrative essence is the intrinsic feature of each atom that, if we know it for every atom in a collection, allows us to best predict the ground truth order of the collection.

\paragraph{Learning Narrative Essence From Data}
\label{sec:learning_narrative_essence_from_data}

If we have a dataset consisting of media collections, we can use it to learn the narrative essence extractor $f_E$.
We can do this with noise contrastive estimation \citep{gutmann2010noise}---specifically, a modification of InfoNCE~\citep{oord2018representation}.
The idea is the following: we give each item in a collection to a learnable feature extractor $f_\theta$, a neural network with parameters $\theta$.
A second learnable function $g_\phi$, a recurrent neural network with parameters $\phi$, takes a sequence $s$ of features as input and produces a scalar score $g_\phi(s)$.
If $g_\phi$ receives a sequence in the correct ground-truth order, $s^* = (f_\theta(x_1), f_\theta(x_2), f_\theta(x_3), ...)$, it should produce a high score.
For randomly ordered sequences, it should produce a low score.
$g_\phi$ can only achieve this if (1) the correct orders of the collections in our dataset have some property that distinguishes them from random orders, and (2) $f_\theta$ learns some atom-wise feature that lets $g_\phi$ recognize this property.
It can be achieved by training feature extractor $f_\theta$ and sequence model $g_\phi$ jointly to minimize the loss
\begin{equation}
    \label{eq:infoNCE_loss}
    \mathcal{L}_\mathrm{N} (\theta, \phi; \mathcal{D}) = -\mathbb{E}_{S \sim \mathcal{D}} \left[ \log \frac{g_\phi(s^*)}{\sum_{s \in S} g_\phi(s)} \right],
\end{equation}
where $\mathcal{D}$ is the dataset of collections with a ground truth order, and $S$ is a set of $N$ sequences that include the correctly ordered sequence $s^*$.
The other $N-1$ sequences in $S$ are random permutations of $s^*$.
The extracted features should be normalized across the sequence so that $g_\phi$ considers the relative, and not the absolute value, of the extracted feature.

In analogy to \citet{oord2018representation}, it can be shown that minimizing $\mathcal{L}_\mathrm{N}$ maximizes a lower bound on the mutual information between the atom-wise features extracted by $f_\theta$ and the order of the collection:
\begin{equation}
    \label{eq:mi_lower_bound}
    I\big(\{f_\theta(x) | x \in c\} ; o(c)\big) \geq \log(N) - \mathcal{L}_\mathrm{N}.
\end{equation}

Additional details are provided in Appendix~\ref{app:narrative_essence_as_mutual_information_with_the_collection_order}.
Formulating narrative essence in the above way enables a general and sound quantitative approach for determining whether collections of a particular type follow any narrative principles as well as what the features of this narrative would be.

\section{Experiments on Music Albums}
\label{sec:experiments_on_music_albums}

In this section, we empirically investigate the concept of narrative essence using the example of music albums.
We selected the FMA dataset~\citep{defferrard2016fma} as it is---at the time of writing---the largest open dataset that includes raw audio files.

While, in principle, highly sophisticated and specialized feature extraction architectures could be used for $f_\theta$, in our experiments, we restrict ourselves to relatively simple and computationally cheap methods.
Each track is represented by features commonly used in music information retrieval that come pre-computed with the available dataset.
These features form a sequence of $75$ vectors of size $7$ (for more details, see Appendix~\ref{app:track_input_features}).
This sequence is the input to the feature extractor $f_\theta$, for which we use a bidirectional LSTM model \citep{hochreiter1997long, graves2005framewise}.
We choose a recurrent feature encoder instead of a feed-forward architecture to give $f_\theta$ more powerful conditional processing abilities.

\begin{wraptable}{r}{0.37\textwidth}
    \centering
    \vspace{-1.38em}
    \caption{Mutual Information (in bits) on the FMA validation set for different dimensionalities of narrative essence. Results are from five runs.}
    \label{tab:mi_for_num_features}
    \begin{tabular}{cc}
        \toprule
        \bfseries Features & \bfseries Mutual Information\\
        \midrule
        $1$ & $1.924 \pm 0.0296$ \\
        $2$ & $1.936 \pm 0.0183$ \\
        $4$ & $1.957 \pm 0.0217$ \\
        $8$ & $1.950 \pm 0.0216$ \\
        $16$ & $1.975 \pm 0.0150$ \\
        \bottomrule
    \end{tabular}
\end{wraptable}

The output of $f_\theta$ is the narrative essence of the given song.
In principle, the narrative essence could be a vector of any size.
However, Table~\ref{tab:mi_for_num_features} shows that a higher dimensional narrative essence leads to only marginal improvements in mutual information captured.
These diminishing returns provide strong evidence that narrative essence, at least for songs in the context of a music album, can be represented as a scalar value.
Note that even a low-dimensional version of narrative essence still captures something much more sophisticated than a basic ranking.
A benefit of using a scalar for narrative essence is that it is directly comparable to other available scalar features (see Section~\ref{sec:narrative_essence_in_comparison_with_other_features}).

We model $g_\phi$ as a bidirectional LSTM as well.
It takes a sequence of narrative essence features as input and computes a scalar score.
In comparison to $f_\theta$, $g_\phi$ has a lower capacity (fewer learnable parameters and more regularization) because there are much fewer full collections (albums) than individual items (songs).
Thus $g_\phi$ is at a considerable risk of overfitting.
A full description of the training setup can be found in Appendix~\ref{app:model_hyperparameters}.
When trained on the full FMA training set\footnote{Note that here and everywhere else we have excluded albums with less than $3$ or more than $20$ tracks.}, the extracted narrative essence achieves a mutual information with the album ordering, as determined by equation \ref{eq:mi_lower_bound}, of ca. $1.924$ bits on the validation set.

\subsection{Narrative Essence in Comparison With Other Features}
\label{sec:narrative_essence_in_comparison_with_other_features}

\begin{figure}[t]
    \centering
    \includegraphics[width=0.45\linewidth]{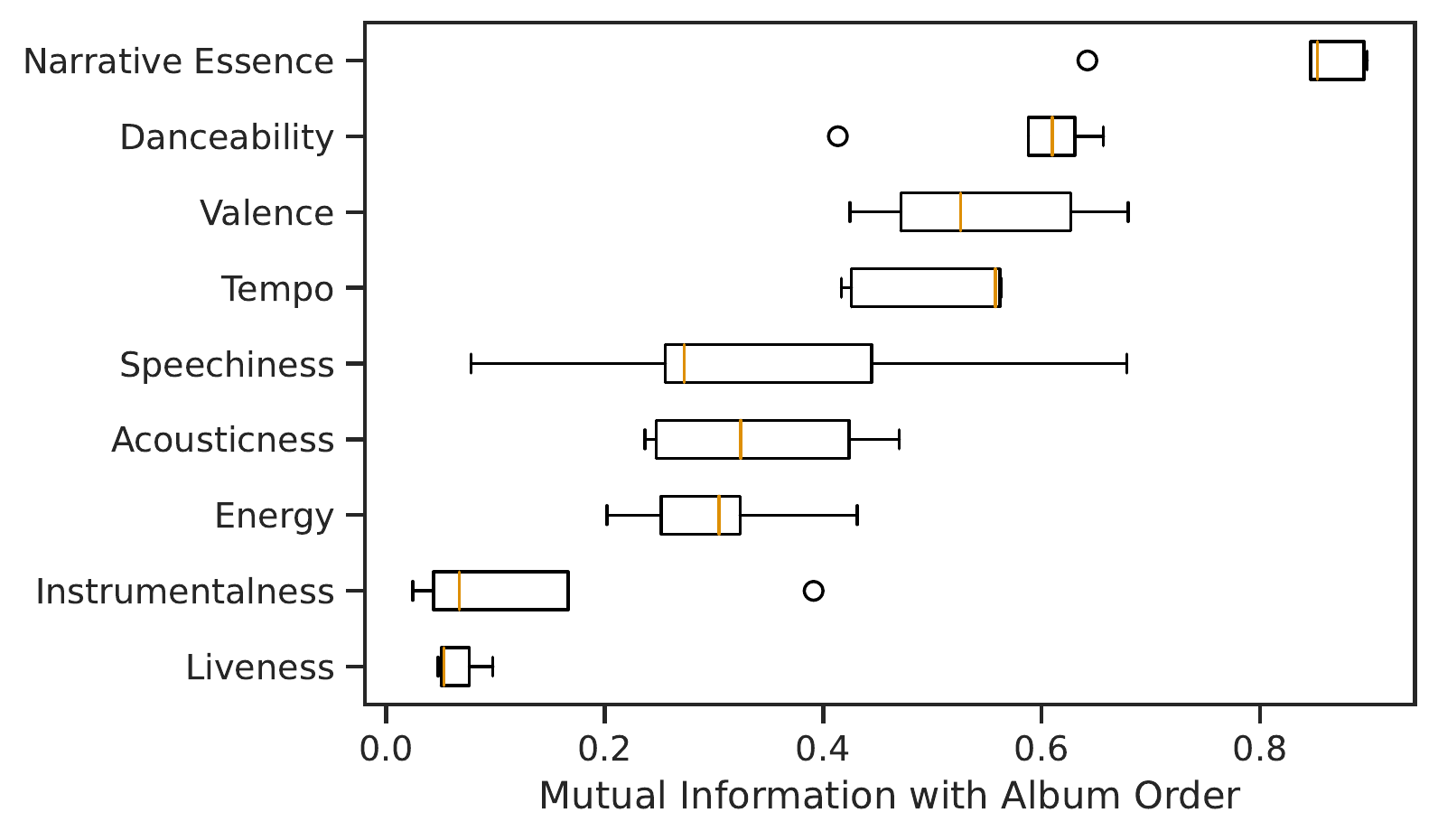}
    \hfill
    \includegraphics[width=0.45\linewidth]{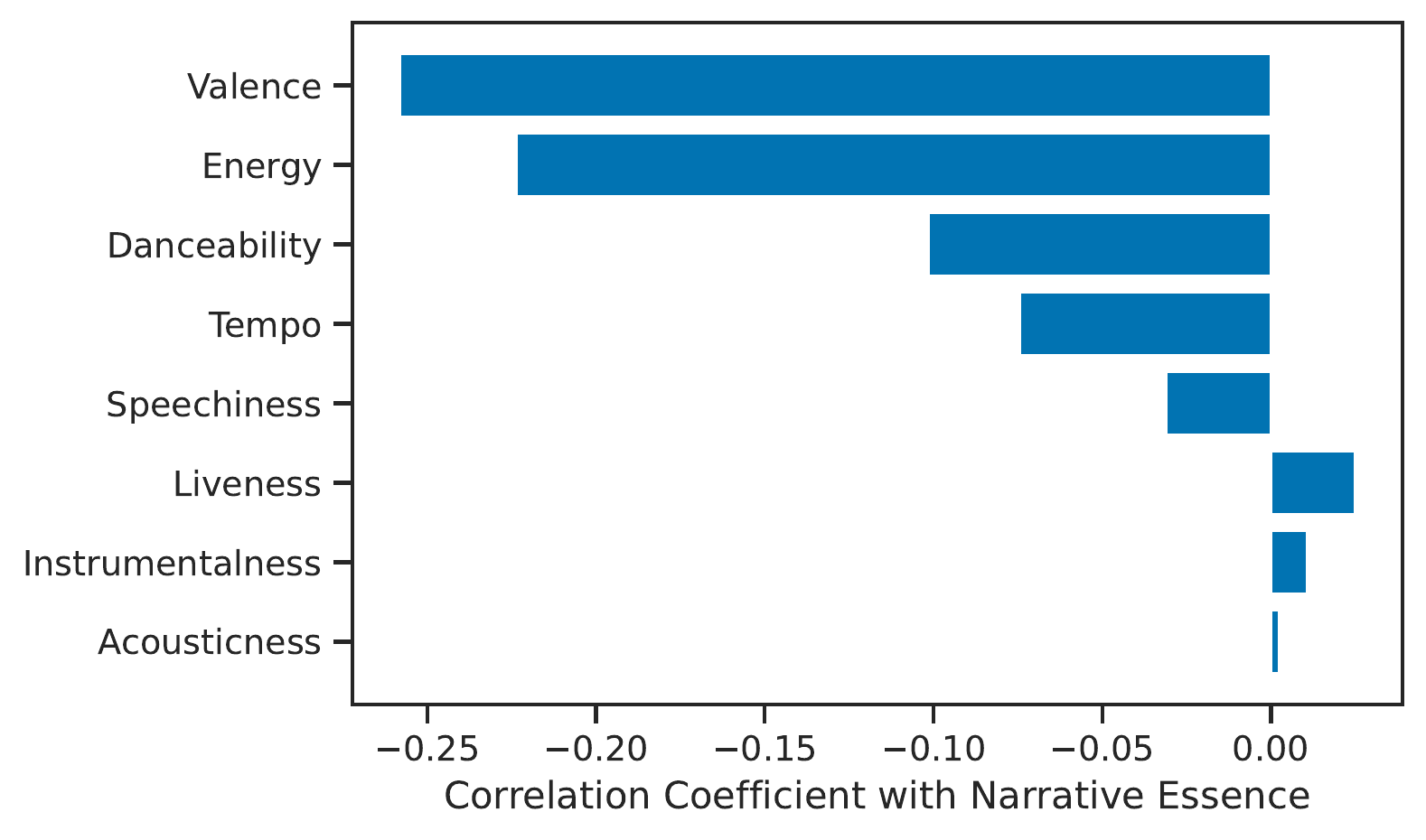}
    \captionof{figure}{\textbf{(left)} The lower bound of the mutual information in bits (see Equation~\ref{eq:mi_lower_bound}) between different features and the album order. Results are shown over five seeds using the validation set. \textbf{(right)} The Pearson correlation coefficient of different features with the narrative essence.}
    \label{fig:feature_correlations}
\end{figure}

When treating the feature extractor $f_\theta$ as fixed and only learning the scoring model $g_\phi$ to minimize $\mathcal{L}_\mathrm{N}$, we can use Equation~\ref{eq:mi_lower_bound} to approximate the mutual information between any available feature and the collection orders.
Here, we compare the narrative essence feature extracted using $f_\theta$ learned on the FMA dataset with some of the other features available in the dataset.
To do so, we learn a dedicated scoring model $g_\phi$ for each available feature---including energy, tempo and valence: a feature designed to capture the mood of a song, roughly ranging from sad to happy~\citep{schubert1999measuring}.
As shown in the left side of Figure~\ref{fig:feature_correlations}, narrative essence has more mutual information with the album order than any of the other features.
Note that the mutual information lower bounds seen in Figure \ref{fig:feature_correlations} are significantly lower than the ones achieved when training on the full dataset (compare Table~\ref{tab:mi_for_num_features}).
This discrepancy is because only a limited subset of the FMA dataset includes the listed pre-computed features when learning the scoring models.
Nevertheless, these results show that our formulation of narrative essence, in combination with the very general processing techniques (i.e., global track feature statistics and an LSTM encoder), robustly outperforms highly engineered features as a candidate for the narrative.

The right side of Figure~\ref{fig:feature_correlations} shows the correlation of the learned narrative essence feature with eight other features.
Here we see a high correlation with the features we expect to be associated (e.g.
valence, energy) and a low correlation with more technical or applied features like acousticness, instrumentalness, and liveness.
Note that the orientation (sign) of the narrative essence is simply a random product of the initialization and has no further meaning; the negative of the narrative essence would have exactly the same amount of mutual information with the collection order.

\subsection{Narrative Essence and Story Templates}
\label{sec:story_templates_and_narrative_essence}

\begin{wrapfigure}{r}{0.4\textwidth}
    \vspace{-4em}
    \begin{center}
        \includegraphics[width=0.4\textwidth]{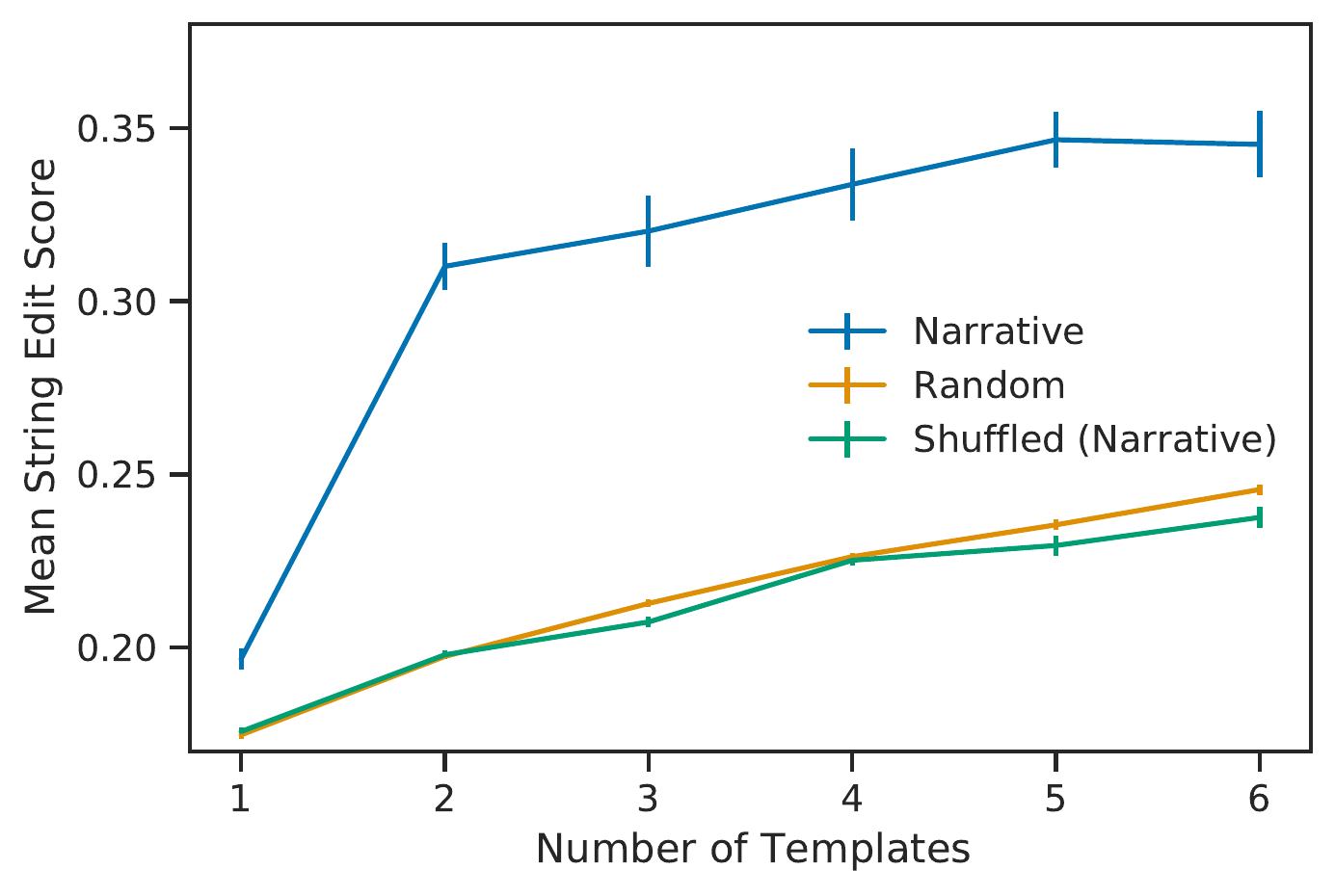}
    \end{center}
    \caption{The performance on the FMA validation set of the templates learned with narrative essence.}
    \label{fig:string_edit_score}
\end{wrapfigure}

The concept of prototypical narratives has been explored in the dramatic arts---such as novels, plays or operas---for a long time \citep{freytag1894technik, vonnegut1981palm, campbell2008hero}.
Other kinds of media, like music albums, may follow different types of narratives.
This section uses the genetic algorithm described in Appendix~\ref{app:story_template_extraction_algorithm} to learn a set of template curves from albums in the FMA dataset using our learned narrative essence feature.

In our algorithm, to evaluate a set of templates, we first fit the album to each template using the novel curve fitting algorithm described in Appendix~\ref{app:template_curve_fitting_algorithm} and then obtain the string edit score\footnote{Preliminary results suggest that similar metrics, such as Spearman's rank correlation coefficient, produce qualitatively similar results.} to determine the similarity of the best fitting to the original order.
We define the string edit score here as $f(T, o) = \max_{\hat{o} \in T}\frac{1}{1 + g(\hat{o}, o)}$ where $T$ is a set of $k$ proposed orders, $o$ is the ground-truth order, and $g$ is the string edit distance (i.e., Levenshtein distance) function.

We compare the performance of the learned templates under narrative essence to two baselines: (1) the maximal mean string edit score of $k$ random orderings and (2) the maximal mean string edit score when the reported narrative essences of the tracks in each album are randomly shuffled.
The latter baseline captures the gain from fitting the narrative essence instead of fitting noise here.
We report these comparisons in Figure~\ref{fig:string_edit_score}.
Our results show with statistical significance ($p < 0.05$) that the album ordering can be partially explained by the narrative essence following our learned templates (details in Appendix~\ref{app:evidence_for_the_existence_of_story_templates}).
See Appendix~\ref{app:fma_template_curves} for a closer look at the template curves found for the FMA dataset and Appendix~\ref{app:demonstration} for a demonstration of these templates in use.
For the source code used to run the above experiments, see Appendix~\ref{app:source_code}.

\section{Related Work}
\label{sec:related_work}

The use of machine learning to derive narrative arcs has previously been explored by \citet{reagan2016emotional}---who used machine learning to explicitly derive the emotional arc of stories from a large corpus of English texts.
More recently, \citet{mathewson2020shaping} used an information-theoretic approach to design a narrative arc and applied this to dialogue generation.
For music playlist ordering, work remains sparse.
However, a considerable body of work has recently emerged in music playlist continuation (e.g., \citet{maillet2009steerable,bonnin2013comparison,vall2019order}).
For a detailed overview of spatial representation of musical form, see \citet{bonds2010spatial}.

\section{Conclusion}
\label{sec:conclusion}

Inspired by their seeming importance in human cognition, this work used machine learning to distill the narrative information from stories.
In doing so, we showed how to produce the narrative essence of their compositional atoms.
We then used evolutionary algorithms alongside narrative essence to extract a series of templates from the music albums in the FMA dataset.
We went on to give statistical evidence that these templates described the ordering of music albums and showed that even one-dimensional narrative essence explains them better than other commonly used music track features.
We hope this work can be applied to partially automate the induction of stories in existing collections of works (e.g., photo galleries, music playlists, etc.).
While we only experimented with music albums here, our work extends to any collection of (largely) independent media.

\section*{Acknowledgements}

We want to thank Kory W. Mathewson and DeepMind Technologies Limited for comments on an earlier version of this work.
This work was supported by the European Research Council (ERC, Advanced Grant Number 742870) and the Swiss National Supercomputing Centre (CSCS, Project s1090).
We also want to thank both the NVIDIA Corporation for donating a DGX-1 as part of the Pioneers of AI Research Award and IBM for donating a Minsky machine.

\printbibliography

\clearpage

\appendix

\section{Narrative Essence as Mutual Information with the Collection Order}
\label{app:narrative_essence_as_mutual_information_with_the_collection_order}

Recall that $c$ is an unordered collection of items $x$, and $o(c)$ is its correct order.
$S$ is a set of $N$ sequences of the encoded items, containing the correct sequence $s^* = (f_\theta(x_1), f_\theta(x_2), f_\theta(x_3), ...)$ (i.e., the one adhering to $o(c)$), and $N-1$ random permutations of $s^*$.
The probability that a particular sequence $s_i$ from $S$ is the correct sequence $s^*$ is
\begin{equation*}
    \begin{split}
        p(s_i = s^* | S, o(c))
        & = \frac{ p(s_i = s^*, S | o(c)) }{\sum_j p(s_j = s^*, S | o(c))} \\
        & = \frac{ p(s_i = s^*) p(S | s_i = s^*, o(c)) }{\sum_j p(s_j = s^*) p(S | s_j = s^*, o(c))} \\
        & = \frac{ p(s_i = s^*) p(s_i | o(c)) \prod_{l \neq i} p(s_l) }{\sum_j p(s_j = s^*) p(s_j | o(c)) \prod_{l \neq j} p(s_l))} \\
        & = \frac{ \frac{1}{N} p(s_i | o(c)) \prod_{l \neq i} p(s_l) }{\sum_j \frac{1}{N} p(s_j | o(c)) \prod_{l \neq j} p(s_l))} \\
        & = \frac{ p(s_i | o(c)) \prod_{l \neq i} p(s_l) }{\sum_j p(s_j | o(c)) \prod_{l \neq j} p(s_l))} \cdot \frac{\prod_k p(s_k)}{\prod_k p(s_k)} \\
        & = \frac{\frac{p(s_i | o(c))}{p(s_i)}}{\sum_j \frac{p(s_j|o(c))}{p(s_j)}}.
    \end{split}
\end{equation*}

With Equation~\ref{eq:infoNCE_loss}, $g_\phi(s)$ is trained to estimate the density ratio $\frac{p(s|o(c))}{p(s)}$.
This means that we can write (following the steps from \cite{oord2018representation})
\begin{equation*}
    \begin{split}
        \mathcal{L}_\mathrm{N}^\mathrm{opt}
        & =  -\mathbb{E}_{S \sim \mathcal{D}} \log \left[ \frac{\frac{p(s^* | o(c))}{p(s^*)}}{\frac{p(s^* | o(c))}{p(s^*)} + \sum_{s \in S_\mathrm{neg}} \frac{p(s|o(c))}{p(s)}} \right] \\
        & \approx \mathbb{E}_{S \sim \mathcal{D}} \log \left[ 1 + \frac{p(s^*)}{p(s^* | o(c))} (N - 1)\right]  \\
        & \geq \mathbb{E}_{S \sim \mathcal{D}} \log \left[ \frac{p(s^*, o(c))}{p(s^*) p(o(c))} N \right] \\
        & = -I(s^*; o(c)) + \log(N) \\
        & = -I(f_E(x_1), f_E(x_2), f_E(x_3), ...; o(c)) + \log(N).
    \end{split}
\end{equation*}

\subsection{Track Input Features}
\label{app:track_input_features}

The FMA dataset provides the following track features: 12 Chroma features, 6 Tonnetz features, 20 MFCC features, Spectral centroid, Spectral bandwidth, 7 Spectral contrast features, Spectral rolloff, RMS energy and Zero-crossing rate.

For every feature, 7 global statistical properties are given: mean, standard deviation, skew, kurtosis, median, minimum and maximum.
We treat these statistical properties as a vector and construct a sequence of these vectors from the 75 features, which constitutes the input for the narrative essence extractor $f_\theta$.
The length of this feature sequence is constant, independent of the track's length.

For tracks that are not included in the FMA data, these features can easily be computed directly from audio standard MIR techniques (the implementation is provided by \cite{defferrard2016fma}).
Before giving the sequence to $g_\phi$, learnable start- and end-of-sequence tokens are added.

\vfill\eject

\subsection{Model Hyperparameters}
\label{app:model_hyperparameters}

All models described in Section~\ref{sec:experiments_on_music_albums} have the same hyperparameters.
The batch size is $16$, $N$ is $32$ (that means we have $31$ negative samples for each example in the batch).

The feature encoder $f_\theta$ is a bidirectional LSTM with $2$ layers, $7$ input features, 128 hidden units and a sigmoid output nonlinearity.
For regularization, we use a dropout of $0.1$ and no weight decay.

The sequence scoring model $g_\phi$ is also bidirectional LSTM with $2$ layers.
It has 32 hidden units and no output nonlinearity.
For regularization, we use no dropout and a weight decay of $10^{-5}$.

For both models, we use the Adam optimizer with a learning rate of $10^{-4}$ and early stopping based on the validation loss.

\vfill\eject

\section{Story Template Extraction Algorithm}
\label{app:story_template_extraction_algorithm}

Extracting a set of narrative arc templates from a collection of albums can be done using Algorithm~\ref{alg:story_template_extraction}.
Note that this algorithm is general and can make use of any collection of media if the narrative essence is replaced by a semantically similar metric.
In our experiments, we always used $\vec{x} = [0.0, 0.2, 0.3, 0.5, 0.65, 0.8, 1.0]^T$.
To derive the value of a template at a given $x$, cubic-spline interpolation is recommended; for the cost of fitting an album to a template, using the mean-squared error is recommended.

\begin{algorithm}
\caption{Story Template Extraction}\label{alg:story_template_extraction}
\hspace*{\algorithmicindent} \textbf{Input:} $\vec{x} = [x_0, x_1, ..., x_q]^T$ where $x_i$ is the relative position of the $i$-th point in the desired templates and a set of albums $\{\vec{a}_1, \vec{a}_2, ..., \vec{a}_n\}$ with each $\vec{a}_i = \{(u_0, v_0), (u_1, v_1), ..., (u_m, v_m)\}^T$ where $u_j$ is the relative position of track $j$ in the album, and $v_j$ is the normalized narrative essence of track $j$\\
\hspace*{\algorithmicindent} \textbf{Output:} set of templates $\{\vec{t}_1, \vec{t}_2, ..., \vec{t}_p\}$ with each $\vec{t}_i = [y_, y_1, ..., y_q]^T$ where $y_j$ is the normalized narrative essence of the $j$-th point in the template\\
\begin{algorithmic}[1]
\State $s \gets$ population size
\State $b \gets$ number of children for each generation
\For{$i \in \{1..s\}$}
    \For{$j \in \{1..p\}$}
        \For{$k \in \{1..q\}$}
            \State $P[i, j, k] \gets \mathcal{N}(0, 1)$
        \EndFor
    \EndFor
\EndFor
\vspace{1em}
\While{not done}
    \State $\sigma = \mathcal{N}(0, 1)$
    \For{$i \in \{1..b\}$}
        \State $father \gets$ random integer in $\{0, 1, ..., s\}$
        \State $mother \gets$ random integer in $\{0, 1, ..., s\} - \{father\}$
        \For{$j \in \{1..p\}$}
            \For{$k \in \{1..q\}$}
                \State $P[s + i, j, k] \gets P[father, j, k]$ with probability $p$ and $P[mother, j, k]$ with probability $1 - p$
                \State $P[s + i, j, k] \gets P[s + i, j, k] + \mathcal{N}(0, \sigma)$
            \EndFor
        \EndFor
    \EndFor
    \For{$i \in \{1..(b + s)\}$}
        \State $\vec{c}_i \gets$ minimum cost as defined in Equation~\ref{eq:ga_cost} for fitting albums using the templates $P[i, :, :]$
    \EndFor
    \State order $P$ in increasing order of corresponding $\vec{c}$
    \State $P \gets P[1:s, :, :]$
\EndWhile
\State \Return $P$
\end{algorithmic}
\end{algorithm}

While many different cost functions for a set of templates could be used here, we use the following:
\begin{equation}
    \label{eq:ga_cost}
    \vec{c} = \sum_{i=1}^n \min_p \frac{1}{l_i} \sum_{j=1}^{l_i} \left( v_i(j) - t_p(j_r) \right)^2,
\end{equation}
where $n$ is the number of albums in the training set, $l_i$ the number of tracks in album $i$, $v_i(j)$ the normalized narrative essence value of the $j$th track of album $i$, and $t_p(j_r)$ is the value of template $p$ at the relative position $j_r = (j - 1) / (l_i - 1)$.
We learn these templates using the training split provided by the FMA dataset and evaluate them on the validation split by fitting the narrative essence of each album to the templates using the algorithm given in Appendix~\ref{app:template_curve_fitting_algorithm}.

\vfill\eject

\section{Template Curve Fitting Algorithm}
\label{app:template_curve_fitting_algorithm}

Deriving an ordering of the media such that their respective values fit a narrative template can be done using Algorithm~\ref{alg:template_curve_fitting}.
The ordering Algorithm~\ref{alg:template_curve_fitting} finds will be minimal first in the maximum deviation of a value from the template curve and minimal second in the average deviation of values from the template curve.
For $n$ items, the worst-case time complexity of this algorithm---provided efficient bipartite matching algorithms such as Hopcroft-Karp \citep{hopcroft1973n} and LAPJVsp \citep{jonker1987shortest} are used---is in \(O(n^{3})\).
In most applications of this work---and for all but the largest collections of independent media---extracting the values that will be fitted will consume vastly more time than the fitting itself.

\begin{algorithm}[h]
\caption{Template Curve Fitting}\label{alg:template_curve_fitting}
\hspace*{\algorithmicindent} \textbf{Input:} normalized values to fit $\vec{y}$ and template curve function $f$ with domain and range $[0, 1]$\\
\hspace*{\algorithmicindent} \textbf{Output:} ordering $\vec{x}$ over values $\vec{y}$ such that the $i$-th value in the ordering is $\vec{x}[i]$
\begin{algorithmic}[1]
\State $\vec{z} \gets \left[f(\frac{0}{|\vec{y}| - 1}), f(\frac{1}{|\vec{y}| - 1}), ..., f(\frac{|\vec{y}| - 1}{|\vec{y}| - 1})\right]^T$
\State $\vec{d} \gets \vec{y} \vec{z}^T$
\vspace{1em}
\State $a \gets 1$
\State $b \gets |\vec{d}|$
\While{$a \neq b$}
    \State $p \gets a + \lfloor (b - a) / 2 \rfloor$
    \State $L, R \gets \{1..|\vec{y}|\}$
    \State $E \gets \{(i \in L, j \in R) \mid \|\vec{y}[i] - \vec{z}[j]\| \leq \vec{d}[p]\}$
    \If{$\exists$ perfect matching for bipartite graph $(L, R, E)$}
        \State $b \gets p$
    \Else
        \State $a \gets p + 1$
    \EndIf
\EndWhile
\vspace{1em}
\State $L, R \gets \{1..|\vec{y}|\}$
\State $E \gets \{(i \in L, j \in R, \|\vec{y}[i] - \vec{z}[j]\|) \mid \|\vec{y}[i] - \vec{z}[j]\| \leq \vec{d}[a]\}$
\State $M \gets$ minimum-cost perfect matching for weighted bipartite graph $(L, R, E)$
\For{$i \in \{1..|\vec{y}|\}$}
    \For{$j \in \{1..|\vec{y}|\}$}
        \If{$(i, j) \in M$}
            \State $\vec{x}[j] = i$
        \EndIf
    \EndFor
\EndFor
\vspace{1em}
\State \Return $\vec{x}$
\end{algorithmic}
\end{algorithm}

\vfill\eject

\section{Evidence for the Existence of Story Templates}
\label{app:evidence_for_the_existence_of_story_templates}

An important question we must address while looking at Figure~\ref{fig:string_edit_score} is whether or not the improvement of the learned curves over the baselines is significant.
This is equivalent to asking if the order of the albums is partially explained by the narrative essence and thus if the narrative structures discovered by our algorithm exist within music albums.
To answer this, we compare the mean string edit score for the selected $k = 4$ templates with the mean string edit score for both baselines on the test set.
We find that the difference observed is significant with a family-wise error rate of $p < 0.05$ using t-tests with Holm-Bonferroni corrections.

\vfill\eject

\section{FMA Template Curves}
\label{app:fma_template_curves}

\begin{figure}[bhpt]
    \centering
    \includegraphics[width=0.7\linewidth]{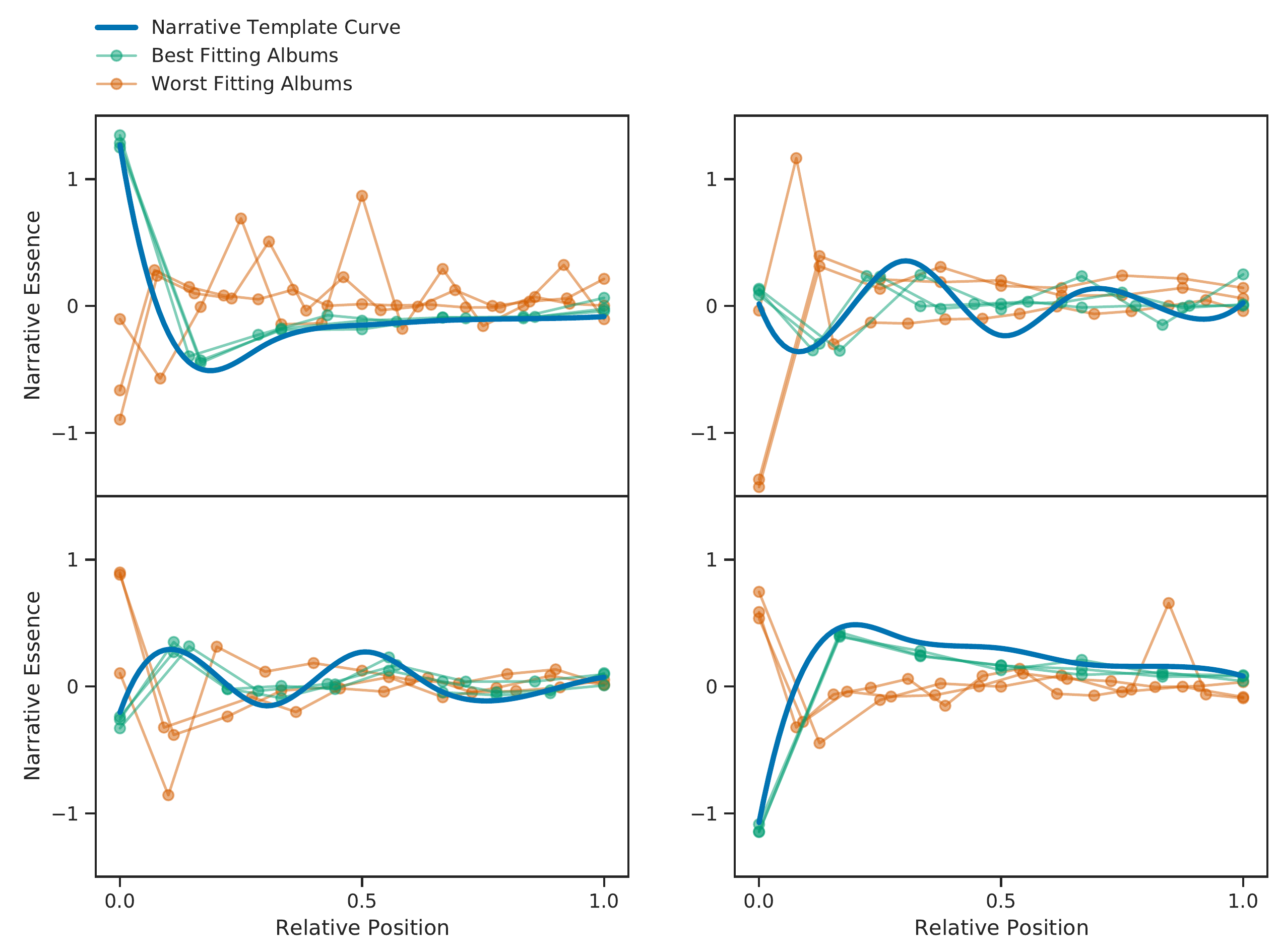}
    \captionof{figure}{Four narrative template curves, as well as the three best and the three worst fitting albums (in terms of mean squared distance) assigned to each curve. Only albums of typical length (between 7 and 15 tracks) are considered in this plot.}
    \label{fig:example_curves}
\end{figure}

During our experiments, we noted that when $k = 3$, a relatively flat curve usually appears in the set of templates.
We thus hypothesize that, similar to when $k = 1$, the presence of this curve is likely a hybrid of many different kinds of story curves.
From $k = 4$ onward, this curve is no longer prominent.
Thus, $k = 4 $ is a good minimal choice for the number of dominant narrative structures in the FMA dataset.
The best-performing learned templates curves for $k = 4$, along with the best and worst fitting albums assigned to these curves, are shown in Figure~\ref{fig:example_curves}.

\begin{figure*}[h]
    \centering
    \includegraphics[width=0.9\linewidth]{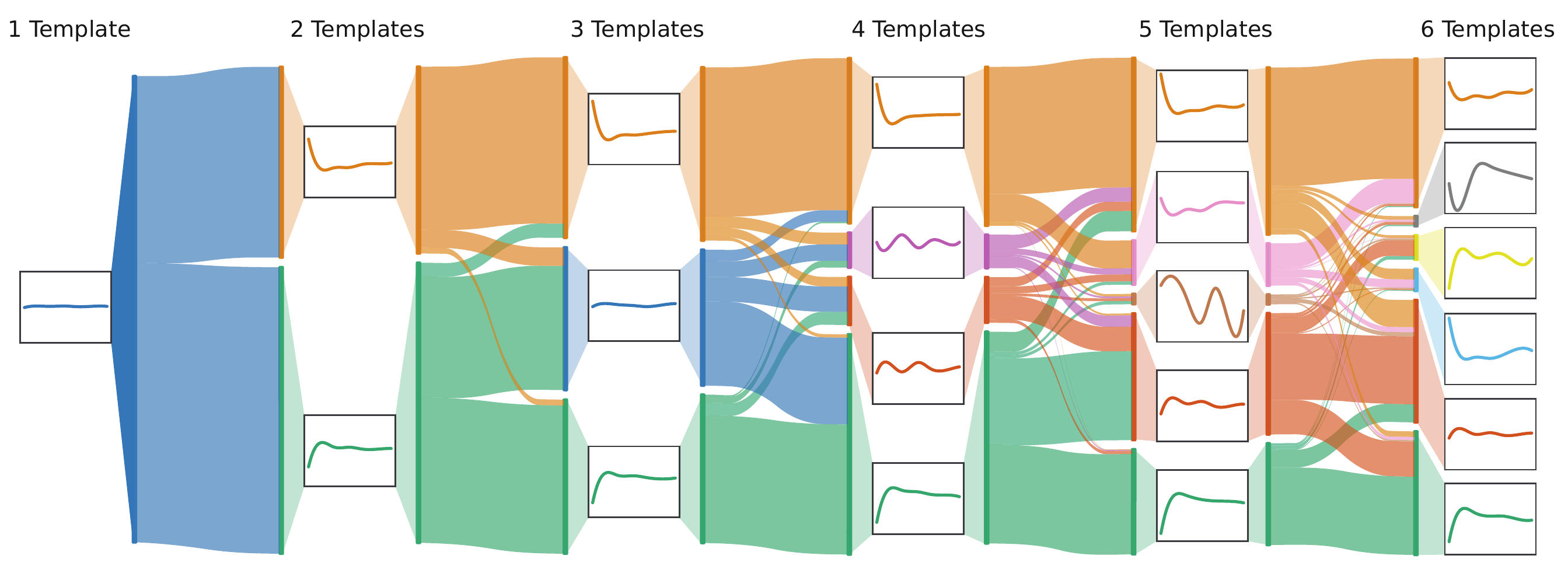}
    \captionof{figure}{How the assignment of individual songs to template curves learned with narrative essence progress as the number of templates increases. Colours show post-hoc analysis to try and find similar prototypical curves.}
    \label{fig:sankey_plot}
\end{figure*}

Figure \ref{fig:sankey_plot} takes a closer look at the narrative templates found in the FMA dataset.
By comparing the learned templates for different $k$, we can draw a genealogy of prototypical narrative arcs---starting from the almost wholly flat average (blue) we for $k = 1$, to an increasing diversity of story curves.

While it is beyond the scope of this work to analyze the genealogy of these curves in detail, we note that the curves here seem to display some notable differences with story curves found elsewhere: where normally there would be a climax, story curves of albums in the FMA dataset seem to instead become more neutral as the album progresses.
Further analysis of this phenomenon is left as future work.

\vfill\eject

\section{Demonstration}
\label{app:demonstration}

\begin{figure}[h!]
    \centering
    \includegraphics[width=0.6\linewidth]{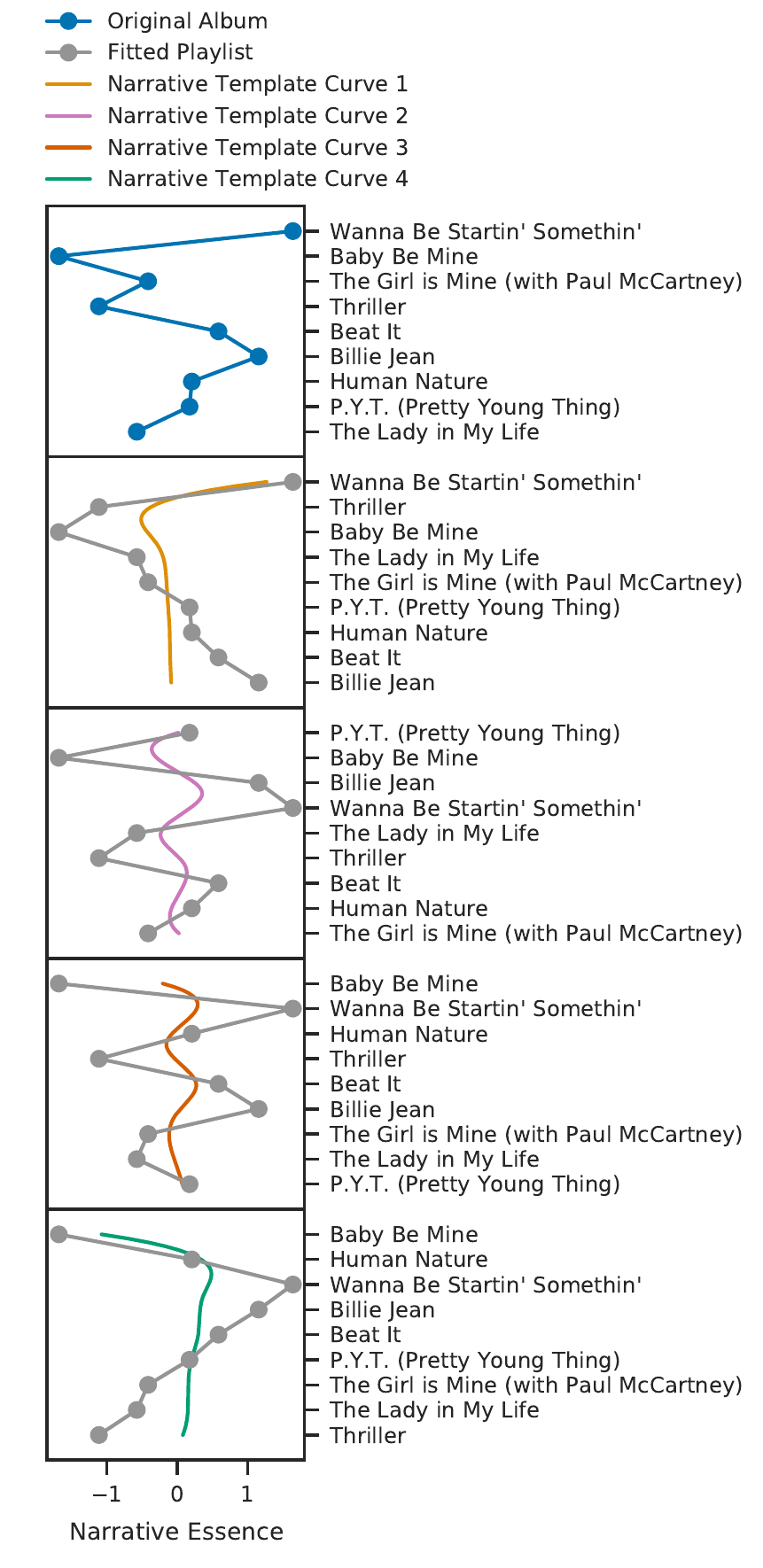}
    \captionof{figure}{Narrative Essence of the album \textit{Thriller} by Michael Jackson---the best-selling original album of all time \citep{riaa2021gold}---in the original order, and fitted to the four narrative template curves found using the method described in Section~\ref{app:story_template_extraction_algorithm}.}
    \label{fig:thriller}
\end{figure}

Figure~\ref{fig:thriller} demonstrates the practical application of this work.
Here, we used the narrative essence feature of the tracks in Michael Jackson's \textit{Thriller} to fit the album to the set of four distinct narrative template curves given in Figure~\ref{fig:sankey_plot}.
By doing so, we have induced several different stories in the album.
The methods presented here can trivially carry out the same task with any album.

\vfill\eject

\section{Source Code}
\label{app:source_code}

The source code used to generate the results presented in this paper is available at \url{https://github.com/dylanashley/story-distiller}

\end{document}